# Robust Extrinsic Self-Calibration of Camera and Solid State LiDAR

Jiahui Liu, *Graduate Student Member, IEEE,* Xingqun Zhan, *Senior Member, IEEE,* Cheng Chi,
Xin Zhang*, *Member, IEEE,* and Chuanrun Zhai

*Abstract*—This letter proposes an extrinsic calibration approach for a pair of monocular camera and prism-spinning solid-state LiDAR. The unique characteristics of the point cloud measured resulting from the flower-like scanning pattern is first disclosed as the *vacant points*, a type of outlier between foreground target and background objects. Unlike existing method using only depth continuous measurements, we use depth discontinuous measurements to retain more valid features and efficiently remove vacant points. The larger number of detected 3D corners thus contain more robust *a priori* information than usual which, together with the 2D corners detected by overlapping cameras and constrained by the proposed circularity and rectangularity rules, produce accurate extrinsic estimates. The algorithm is evaluated with real field experiments adopting both qualitative and quantitative performance criteria, and found to be superior to existing algorithms. The code is available on GitHub[1].

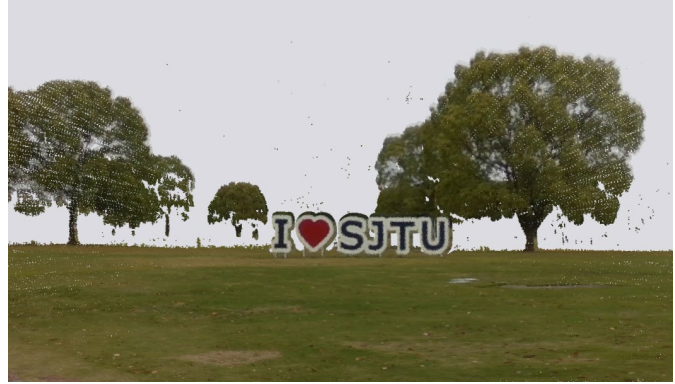

Figure 1. Colorized point cloud visualization using the estimated extrinsic parameters from our approach. The uncolored point cloud lacks pixel information, owing to the devices relative position and field-of-view (FoV).

## I. Introduction

POINT cloud, a data format usually generated by Light Detection and Ranging (LiDAR), together with camera images, are commonly fused in autonomous driving including tasks such as object classification [1, 2], SLAM [3, 4], and navigation [5]. Visual information contains rich representation of the surroundings including color, texture, and shapes, while laser point cloud is characterized by its accurate 3D geometric information regardless of illumination conditions. The complementary properties of these two sensors will evidently enhance the robustness of the final fused results. But before that, an extrinsic calibration step is inevitable. Existing methods examined varied types of geometric correspondence, fisheye vs pinhole cameras, 2D laser range finder (LRF) vs 3D LiDAR. Nevertheless, with the rapid development of the point cloud devices concerning the resolution, field of view (FoV), and scanning pattern [6], the solid-state LiDAR stands out as a burgeoning cost-effective device, and this paper is dedicated to the extrinsic calibration of this new type of LiDAR with cameras, exploiting the LiDAR's unique unevenly distributed flower-like scanning pattern. With this defining feature, the point cloud density of the studied sensor can easily exceed that of a mainstream 64-line mechanical rotating LiDAR, and the non-repetitive scanning is the bionic design of the retina, making the center, usually the region of interest (ROI), possess the highest density.

A new challenge emerges in the joint calibration due to this feature of LiDAR: the unevenly dense point cloud of ROI brings considerable outliers resulting from the scanning resolution, distance, and object properties that invalidate conventional approaches. Besides, for image features, circularity and rectangularity embedded in the calibration pattern are not fully utilized. Despite recent developments [7], current approaches overlook these factors.

In this letter, we propose an automatic extrinsic calibration approach applicable to solid-state LiDAR adopting the target-based methodology, especially for open field scenes characterized by lack of features. The corner features are precisely extracted due to obtained *a priori* board information, as opposed to the targetless method which might fail in field scenes. To overcome these above drawbacks and pursue a high accuracy result commensurate with the high-resolution LiDAR, we solve two unique target-based calibration issues of the solid-state LiDAR and camera: 1). We use voxelization and depth-discontinuous points to eliminate the outliers, and 2). We design a special calibration board and propose two corresponding pixel constraints to assist feature extraction of the camera. Specifically, the contributions of our work are as follows:

This work was supported by the National Natural Science Foundation of China under Grant 62173227, and the National Natural Science Foundation of China under Grant 62103274. (*Corresponding author: Xin Zhang.)*

Jiahui Liu, Xingqun Zhan, Cheng Chi, and Xin Zhang are with the School of Aeronautics and Astronautics, Shanghai Jiao Tong University, Shanghai 200240, China (e-mail: jh.liu@sjtu.edu.cn; xqzhan@sjtu.edu.cn; chichengcn@sjtu.edu.cn; xin.zhang@sjtu.edu.cn).

Chuanrun Zhai is with the ComNav Technology Ltd., Shanghai 201801, China (e-mail: zhaichuanrun@sinognss.com).

[1]https://github.com/GAfieldCN/automatic-camera-pointcloud-calibration

- We re-examine the causes and characteristics of the *vacant points* (outliers between the foreground and background objects) featured in solid-state LiDAR, and propose a removal procedure by voxelization and depth-discontinuous points. Compared to the state-of-the-art method that uses depth-continuous points, our method tends to retain more line features after outlier detection and removal, in open field scenes.

- We design a tailored calibration board conducive to point cloud outlier removal, and also recognizable for cameras by applying new pixel constraints called *rectangularity* and *circularity* for effective image feature extraction.

- We develop a companion calibration software and open source it on Github[1].

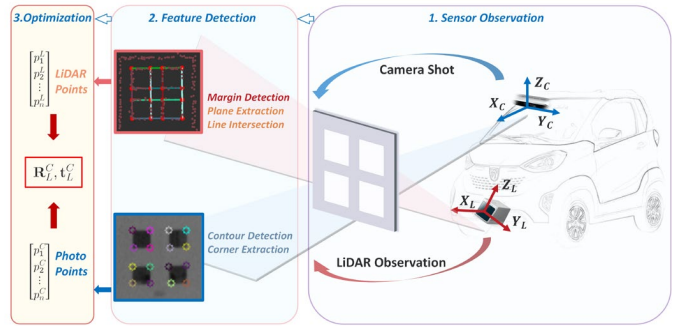

Figure 2. The pipeline of the proposed approach. The calibration target is a board with four square holes observed by both of the sensors. After (I) detecting planes and obtaining the intersection of lines from the point cloud, (II) discerning contours and extracting corners of each contour from the image, the extrinsic parameters could be reached by aligning corresponding selected points and solving an optimization problem.

## II. RELATED WORK

The extrinsic calibration between camera and LiDAR has been well-studied in robotics, spurred by information fusion of multimodal sensing systems. Manual feature selection [8] is naturally considered a generic approach, which extracts distinct corner points by each sensor from natural or artificial scenes, but is often cumbersome to use since the complicated procedures and the induced mismatched errors are almost unacceptable. In contrast, there is a class of more convenient and often called automatic calibration approaches that are capable of extricating human operators from the onerous work of selecting features. Automatic calibration normally falls into two categories: online and offline methods [9], depending on whether artificial targets are adopted. The online techniques (thus must be targetless approaches) detect general planar or line features embedded in the surroundings to estimate the extrinsic parameters, while the offline (target-based) techniques rely on manmade geometries crafted for feature extraction, thus admitting more accurate and robust results compared with the online approaches.

The online approaches are usually considered in dynamically changing environments or sensor platforms requiring frequent installation and removal from a test vehicle. Checkerboards were commonly applied [10] [11] in offline calibration methods where they provided planar constraints for LRFs. By extracting corners in the image and the corresponding pose in the point cloud, the extrinsic parameters could be obtained by minimizing the reprojection error, but the required 5 poses might be burdensome. Consequently, G. Koo *et al.* [12] propose analytically derived covariances of checkerboard planes for convenient and precise calibration of mechanical spinning LiDAR, and line-plane [13], point-plane [14] correspondences are utilized to enhance the robustness of the approach. However, the checkerboard patterns might pollute point clouds due to their black and white colors [15], thus triangular boards [16] and circular targets [17] serve as planar constraints to achieve geometric minimization by fitting selected points. While planar boards are frequently considered, several researchers turn to spatially geometric objects for calibration since the unambiguous features will facilitate accurate corner association to some extent. Z. Pusztai *et al.* [15] use boxes to calibrate since the intersections of their planes could be easily attained and rightfully constrained due to the perpendicular and square faces, and T. Tóth *et al.* [18] reveal how the spheres can be estimated if their contours are detected. C. Guindel *et al.* [19] have designed a unique calibration board with four circular holes.

The second branch of automatic calibration approaches adopts the features in natural scenes without any deliberately prepared targets. P. Moghadam *et al.* [20] exploit natural linear features extracted from both 3D point clouds and 2D images, eluding the hurdle of modifying the scene with artificial targets. J. Rehder *et al.* [21] apply a RANSAC algorithm to detect planes. Due to the limitations of the two above methods concerning whether the geometrical features are sufficient for robust calibration, C. Park *et. al.* [9] propose a structureless approach combining a closed-form solution with a modified bundle adjustment. Meanwhile, J. Jeong *et. al.* [22] present a calibration method for non-overlapping cameras and LiDAR by capturing the informative road markings, X. Zhang *et. al.* [23] select the line feature for its ubiquity to improve universality. For the solid-state LiDAR, an approach [24] tailored for high-resolution LiDAR aligns natural edge features from different sensory modalities to complete the automatic targetless calibration process, and J. Cui *et. al.* [25] provide a fully automatic approach adopting the checkerboards.

Our proposed method is a target-based method. Compared to [11] and [25], our method efficiently eliminates the adverse impact of the outliers and outperforms the others on average; compared to the targetless method [24], our method is efficient in outdoor environments where lacks of features is the biggest challenge for LiDAR measurement.

## III. CALIBRATION APPROACH

### A. Overview

The calibration approach is intended to estimate the rigid body transformation $\mathbf{T}_L^C \in SE(3)$ between the LiDAR frame $L$, whose origin is the point cloud measurement center defined

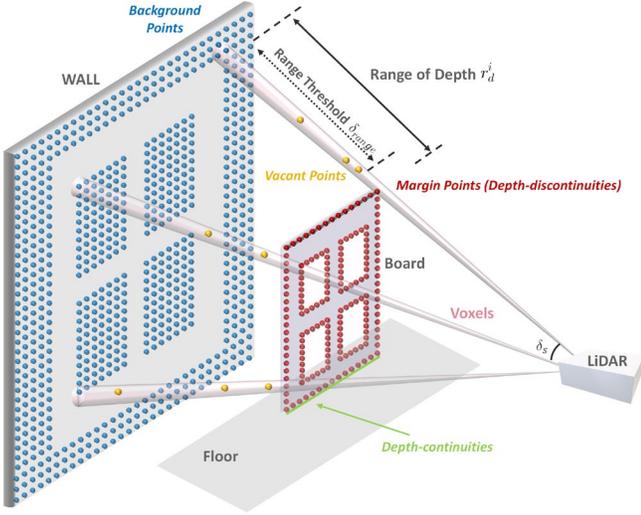

Figure 3. The margin points detection process. The points are grouped by small voxels, and if the range of depth is greater than the threshold, only the point nearest to the LiDAR would be regarded as the margin points.

by the device manufacturer, and the camera frame $C$, whose origin lies on the optical center, denoted as

$$\mathbf{T}_L^C := \begin{bmatrix} \mathbf{R}_L^C & \mathbf{t}_L^C \\ \mathbf{0}^T & 1 \end{bmatrix} \in \mathbb{R}^{4\times 4} \quad (1)$$

where $\mathbf{R}_L^C \in SO(3)$ and $\mathbf{t}_L^C \in \mathbb{R}^3$ that represents the rotation matrix and the translation vector, respectively.

Figure 2 illustrates the overall process of the proposed method. A planar target with four square holes is designed due to the easily distinguished characteristics, and it should be perceived by both of the sensors to obtain the accumulated data. Then the proposed feature extraction algorithm would be applied to produce the corner points of each square as features in the 2D image and 3D point cloud, which are aligned and fed into the optimization process to solve a PnP (Perspective-n-Point) problem.

*B. Point Cloud Feature Extraction*

*1) Depth-discontinuous Margin Points Detection:* Since the raw accumulated point cloud data is extremely massive, the first step of our method is to select the points containing valid information about the calibration board, called *margin points*, and screen out the meaningless ones, thus conducive to the decrease of the following computational complexity.

Some existing work [24] extracts the depth-continuities, defined as the plane-joining lines with continuous depth, as shown in Figure 3. Though the outliers are removed, many valid features are erased as well (cf. Figure 7 of [24]), causing "bad scenes" where lack of sufficient edge features is the biggest challenge to complete the calibration process, and this is even worse in open field scenes than half occluded or pure indoor settings. Some researchers [19] start the algorithm by identifying the depth-continuities by assigning each point a magnitude representing the depth difference with other points. These methods might work well for mechanical spinning LiDAR, but after thorough research about the laser measurement principle and through our practice of the solid-state LiDAR, we found that the detected depth-discontinuous points might fall within the vacant space, called the *vacant points* as illustrated in Figure 3. This phenomenon is caused by the fake connections of the adjacent laser beam reflected by different materials, and will get the above method into trouble since it would mistakenly assume the vacant points as the depth discontinuities. In response, we propose a method denoted as Algorithm 1, and the performance is illustrated in Figure 4. Meanwhile, the experimental values of the threshold (*margin step* $\delta_s$, range threshold $\delta_{range}$, plane threshold $\delta_d^{Plane}$, line threshold $\delta_d^{Line}$) are described in Section IV.

**Algorithm 1** Point cloud feature extraction
**Input:** Raw LiDAR points $\mathbf{P}$
**Output:** LiDAR corner points $\{\hat{\mathbf{P}}^{Lcp}\}$
1:   **for** Points $\{\mathbf{P}_i\}$ in each voxel $i$ **do**
2:     $\{\mathbf{P}^F\} \leftarrow meanshift(\mathbf{P}_i)$
3:     Compute range of depth $r_d^i$
4:     **if** $r_d^i > \delta_{range}$ **then**
5:       $\{\mathbf{P}^L\} \leftarrow deflate(\mathbf{P}^F)$
6:   **while** not enough $\{\mathbf{P}^{Plane}\}$ **do**
7:     $(\mathbf{P}_a^L, \mathbf{P}_b^L, \mathbf{P}_c^L) \leftarrow randomselect(\mathbf{P}^L)$
8:     $(\mathbf{n}_i^L, d_i^L) \leftarrow getplane(\mathbf{P}_a^L, \mathbf{P}_b^L, \mathbf{P}_c^L)$
9:     **for** $\mathbf{P}_k \in \{\mathbf{P}^L\}$ **do**
10:       $\{\mathbf{P}_i^{Plane}\} \leftarrow \mathbf{P}_k$ if $d_i^{Plane} < \delta_d^{Plane}$
11:     $(\mathbf{P}_i^{Plane}) \leftarrow selectbest(\mathbf{P}_i^{Plane})$
12:     remove $\{\mathbf{P}_i^{Plane}\}$ from $\{\mathbf{P}^L\}$
13:   **while** not enough $\{\mathbf{P}^{Line}\}$ **do**
14:     $(\mathbf{P}_a^L, \mathbf{v}_i) \leftarrow randomselect(\mathbf{P}^{Plane})$
15:     **for** $\mathbf{P}_k \in \{\mathbf{P}^{Plane}\}$ **do**
16:       $\{\mathbf{P}_i^{Line}\} \leftarrow \mathbf{P}_k$ if $d_i^{Line} < \delta_d^{Line}$
17:     $(\mathbf{P}_i^{Line}, \mathbf{P}_a^L, \mathbf{v}_i) \leftarrow selectbest(\mathbf{P}_i^{Line})$
18:     remove $\{\mathbf{P}_i^{Line}\}$ from $\{\mathbf{P}^{Plane}\}$
19:   $(\mathbf{v}_{hor}^n, \mathbf{v}_{ver}^n) \leftarrow weight(\mathbf{v}_i, \{\mathbf{P}_i^{Line}\})$
20:   $\{\mathbf{P}_{a\ hor}^L, \mathbf{P}_{a\ ver}^L\} \leftarrow list(\{(\mathbf{P}_a^L, \mathbf{v}_i)\}$
21:   $\{\hat{\mathbf{P}}^{Lcp}\} \leftarrow intersect(\mathbf{v}_{hor}^n, \mathbf{v}_{ver}^n, \{\mathbf{P}_{a\ hor}^L, \mathbf{P}_{a\ ver}^L\})$

In our algorithm, we separate the laser scanning space into myriad small voxels according to the elevation and azimuth angle spanning the device's entire FoV, and extract the margin points by three consecutive steps: 1) The points in a forward-located range within each detecting voxel are firstly clustered using algorithm Meanshift, and would converge to a set of points with the highest density while in the foreground, called the *frontier points* $\mathbf{P}^F$, so most of the vacant points could be erased since they usually hide behind the frontiers. 2) We deflate the set of frontier points by including sensor measurements within one standard deviation region around the frontier points, since the range measurements error would cause the foreground objects inflation, causing biased frontier detection. The one standard deviation threshold depends on integrity-based rules

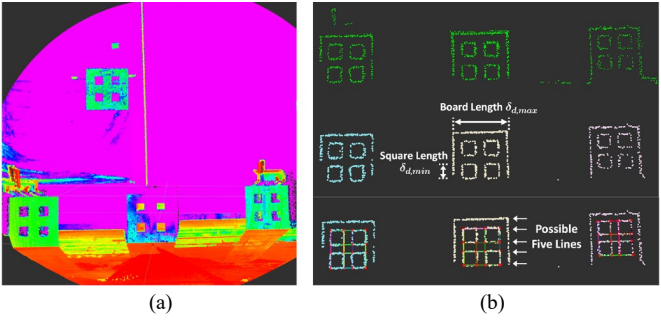

Figure 4. (a) Raw point cloud data; (b) Margin points, detected planes, and extracted corners (lines intersections, the red points) from top to bottom.

[26], which bound the measurements with a cumulative distribution function. 3) For each voxel iterated by the *margin step* $\delta_s$, a *range of depth* $r_d^i$ is calculated, considering a threshold $\delta_{range}$, the margin points could be extracted with $r_d^i > \delta_{range}$, otherwise the frontiers belong to the *background points*, which means they are the depth-continuous points, e.g., the wall surface. The overall process is visualized in Figure 3.

Through the above process, only the margin points representing the contours of depth-variant objects are extracted, which could reduce the computational workload of the following plane detection algorithm. Yet some noise points are still not removed, thus we conduct an optional filter process that assigns each margin point a descriptor $N$ describing the degree of sparseness, about the number of points whose Euclidean distance $d$ to each margin point is smaller than the threshold $\delta_N$, and eliminating the point that $N < \delta_N$.

  2) *Plane Detection:* The margin points roughly depict the contours of the target, and we still need to separate the planes involved in the target from the point cloud. A RANSAC algorithm is applied for plane detection, which selects a cluster of points with the maximum number of inliers.

In each iteration, three non-collinear points are randomly chosen based on the 3D plane equation to form one plane

$$\mathbf{n}_i^L \mathbf{P}^L + d_i^L = 0 \quad (2)$$

where the $\mathbf{n}_i^L \in \mathbb{R}^3$ is the 3D normal vector of the plane, with $d_i^L$ as the distance bias. The Euclidean distance of any point $\mathbf{P}_i^{Plane}$ to the plane $d_i^{Plane}$ is calculated, and if it is smaller than the threshold $\delta_d^{Plane}$, the point will be merged into the cluster,

$$d_i^{Plane} = \frac{\mathbf{n}_i^L \mathbf{v}^L}{\|\mathbf{n}_i^L\|_2} < \delta_d^{Plane} \quad (3)$$

where $\mathbf{v}^L$ is the vector from the point $\mathbf{P}_i^{Plane}$ to any point on the plane.

After the extraction of the dominant plane, the cluster would be saved and removed from the point cloud, and the RANSAC process will continue until all planes are considered detected. The whole process could mostly erase the points not stemming from the calibration board, and also serve as a filter controlling the number of detected planes with the adjustment of $\delta_d^{Plane}$.

  3) *Line Detection and Corner Extraction:* For the current target, corner features are intersections of perpendicular lines, thus another RANSAC algorithm is applied for 3D line detection. Similarly, two randomly chosen points could determine a line with one point $\mathbf{P}_a^L$ and a direction vector $\mathbf{v}_i$, thus the points $\mathbf{P}_i^{Line}$ are clustered with

$$d_i^{Line} = \frac{(\mathbf{P}_a^L - \mathbf{P}_i^{Line}) \times \mathbf{v}_i}{\|\mathbf{v}_i\|_2} < \delta_d^{Line} \quad (4)$$

**Algorithm 2** Image feature extraction
**Input:** Image $img$, Intrinsic $\boldsymbol{K}$
**Output:** Camera corner points $\{\hat{\mathbf{P}}^{Ccp}\}$
1: $img_{rect} \leftarrow rectify(img, \boldsymbol{K})$
2: $contours_{all} \leftarrow detectcontour(img_{rect})$
3: $contours_{square} \leftarrow contourfilter(contours_{all})$
4: **for** each *pixel point* in each contour $i$ **do**
5: $\quad (\mathbf{P}_a^C) \leftarrow randomselect(contours_{square}^i)$
6: $\quad (\mathbf{P}_1^C) \leftarrow Maxdistance2(\mathbf{P}_a^C)$
7: $\quad (\mathbf{P}_2^C) \leftarrow Maxdistance2(\mathbf{P}_1^C)$
8: $\quad (\mathbf{P}_3^C) \leftarrow Maxdistance2(\mathbf{P}_1^C, \mathbf{P}_2^C)$
9: $\quad (\mathbf{P}_4^C) \leftarrow Maxdistance2(\mathbf{P}_3^C)$
10: $\quad \{\mathbf{P}^C\} \leftarrow list(\mathbf{P}_1^C, \mathbf{P}_2^C, \mathbf{P}_3^C, \mathbf{P}_4^C)$
11: $\{\hat{\mathbf{P}}^{Ccp}\} \leftarrow list\{\mathbf{P}^C\}$

Each extracted line will then be labeled as either horizontal or vertical according to their direction vector $\mathbf{v}_i$, but since the lines in each group might not strictly be parallel to the others, we conduct a fitting procedure by assigning each vector $\mathbf{v}_i$, a weight depending on the ratio of the number of points in the line to the total number of points in all the considered lines, and obtaining the horizontal and vertical clustered vector $\mathbf{v}_{hor}^n, \mathbf{v}_{ver}^n$. Take the horizontal one as an example:

$$\mathbf{v}_{hor}^n = \sum_{i=1}^n \mathbf{v}_i \cdot \frac{n_i}{n_{hor}^{total}} \quad (5)$$

where $n_i$ represents the number of points contained in each line $\mathbf{P}_i^{Line}$, and $n_{hor}^{total}$ is the total number of points of all horizontal lines. This procedure could improve the precision of our algorithm, since some mistakenly extracted lines with $\mathbf{v}_i$ deviating from the $\mathbf{v}_{hor}^n$ would contribute trivially to the normalized vector because they contain fewer points compared with real borderlines.

To erase some interfering lines, the distances between every two lines are calculated, and 1) if the distance is approximately equal to the length of the board $\delta_{d,max}$, both of the lines will be removed, i.e., they are the two edges of the board; 2) if the distance is smaller than the length of the edge of the squares $\delta_{d,min}$, the line containing fewer points will be eliminated, i.e., it is the mis-extracted line that coincides with the other. Especially, only one edge of the board will be observed by the LiDAR under some special circumstances (e.g., Figure 4 (b), the lower edge of the board is 'missing' when the board is placed on the ground), thus in the five horizontal or vertical lines, the one with the maximum points would be erased since it is the only remaining edge. Also note that the points $\mathbf{P}_i^{Line}$ and the vector $\mathbf{v}_{hor}^n$ or $\mathbf{v}_{ver}^n$ might neither reside in nor be parallel to the plane, thus they have to be projected onto the plane and the projected lines will be used for the next processing stages.

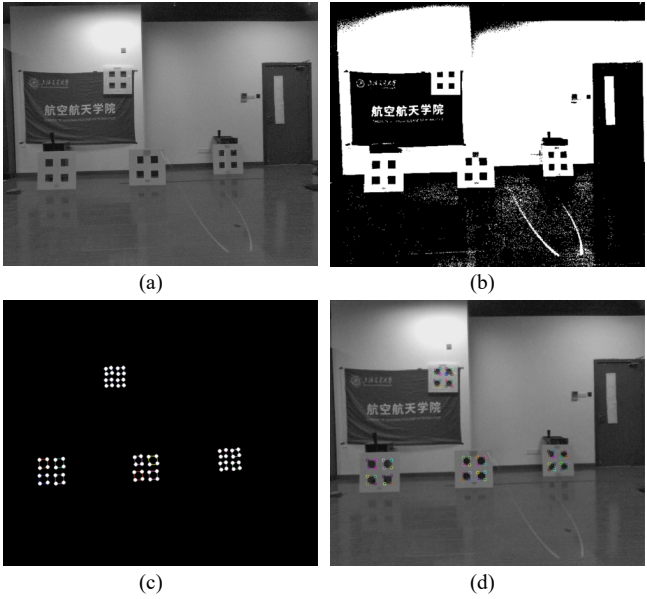

Figure 5. (a) Rectified raw image; (b) Binary pictures; (c) Contour detection and corner extraction; (d) Precise sorted corners.

Therefore, the intersections of all horizontal lines and vertical lines are the point features, as illustrated by Figure 4 (b).

### C. Camera Image Feature Extraction

The intrinsic parameters including the camera focus and the distortion parameters, denoted as $\boldsymbol{K}$, are assumed known *a priori*, since the camera usually needs to be pre-calibrated in most cases. So the imported raw image is firstly rectified to obtain the undistorted picture. Then we could obtain the camera image features via contour detection and corner extraction, as illustrated in Algorithm 2.

*1) Contour Detection:* This is a broadly-studied problem with plenty of state-of-the-art approaches. Our work extracts the topological structure of the binary pictures for border following, which is also the algorithm encapsulated in OpenCV. The contours of the surroundings could be detected by discriminating between the hole borders and the outer borders. To catch the expected square contours, a range of the contour area is initially set to screen out the unreasonably large or small contour objects. We define two simple metrics of *rectangularity* and *circularity* to describe the geometric characteristic of a detected/expected contour, thus ensuring the robustness of extracted image features. Rectangularity is a parameter representing the similarity between its contour and a rectangle, reflecting the degree of saturation of the contour to its minimum enclosing rectangle, while circularity represents the similarity between a contour to a circle, and serves as a restriction to guarantee the square shape and screen out the long slender rectangles.

$$rectangularity = S_{MER} \;/\; S_{Hull} \qquad (6)$$

$$circularity = 4\pi \cdot S_{Area} \;/\; length^2 \qquad (7)$$

where the $S_{MER}$ is the pixel area of the minimum enclosing rectangle, $S_{Hull}$ is the area of the convex hull, $S_{Area}$ represents the pixels encompassed by the contour, $length$ is the perimeter of the contour.

*2) Corner Extraction:* Each of the detected contours contains dozens of pixel points, and we are only interested in the four corners. The 2D Euclidean distance of one randomly selected initial point and the rest of the points is calculated, and the point with the largest distance is regarded as the first corner. This procedure concerning the first corner and others is repeated to get the second corner, lying on one of the two diagonals with the first corner. The third one could be obtained by measuring the sum of the distance to the first and the second, while the fourth corner is again on the opposite end of the second diagonal. Please note that the four corners should be arranged in a specific order, e.g., the upper-left is the first to benefit the feature registration process. The outputs at each stage of the procedure are shown in Figure 5.

### D. Feature Registration and Optimization Process

The extracted features should be arranged in a specific order before the optimization. The point cloud features $\mathbf{P}_i^L = [X_i, Y_i, Z_i]^T$ and the image features $\mathbf{P}_i^C = [u_i, v_i, 1]^T$ are sorted according to their relative positions in each board, and satisfy the following relation

$$s \cdot [u_i, v_i, 1]^T = \boldsymbol{K}(\mathbf{R}_L^C [X_i, Y_i, Z_i]^T + \mathbf{t}_L^C) \qquad (8)$$

where $\mathbf{R}_L^C$ is an orthonormal matrix, $\mathbf{t}_L^C$ is the translation vector, $s$ is the scale factor, $\boldsymbol{K}$ is the camera intrinsics matrix,

$$\boldsymbol{K} = \begin{bmatrix} f_x & 0 & c_x \\ 0 & f_y & c_y \\ 0 & 0 & 1 \end{bmatrix} \qquad (9)$$

Then we could solve for the extrinsics from the following nonlinear problem

$$\begin{aligned}
& \arg\min_{\hat{\mathbf{R}}_L^C, \hat{\mathbf{t}}_L^C} \frac{1}{2} \sum_{i=1}^n \|\mathbf{P}_i^C - \frac{1}{s}\boldsymbol{K}(\mathbf{R}_L^C \mathbf{P}_i^L + \mathbf{t}_L^C)\|^2 \\
& = \arg\min_{\hat{\mathbf{T}}_L^C} \frac{1}{2} \sum_{i=1}^n \|\mathbf{P}_i^C - \pi(\mathbf{T}_L^C \mathbf{P}_i^L)\|^2
\end{aligned} \qquad (10)$$

where $\|\cdot\|^2$ denotes the $L_2$ norm of a vector, $\pi(\cdot)$ is the pin-hole projection model.

For a monocular camera, this process is a PnP problem that could be solved in an iterative way, which requires the derivative of the tangent space of the transformation, i.e., the Jacobian $\mathbf{J_T}$. Let $\boldsymbol{\xi} = [\boldsymbol{\rho} \; \boldsymbol{\phi}]^T \in \mathfrak{se}(3)$ be the Lie algebra element corresponding to the Lie group $\mathbf{T}_L^C$ that could be transferred by the exponential map

$$\mathrm{Exp}(\boldsymbol{\xi}) = \begin{bmatrix} \exp([\boldsymbol{\phi}]_\times) & \boldsymbol{\rho} \\ \mathbf{0}^T & 1 \end{bmatrix} \qquad (11)$$

where $[\boldsymbol{\phi}]_\times \in \mathfrak{so}(3)$ represents a skew-symmetric matrix. To generalize the concept of the derivative, perturbations to $\boldsymbol{\xi}$ in $\mathfrak{se}(3)$, $\delta\boldsymbol{\xi}$, can be done as

$$\begin{aligned}
\mathrm{Exp}(\delta\boldsymbol{\xi} \oplus \boldsymbol{\xi}) &= \exp([\delta\boldsymbol{\xi}]_\times) \exp([\boldsymbol{\xi}]_\times) \\
&\approx (\mathbf{I} + [\delta\boldsymbol{\xi}]_\times) \exp([\boldsymbol{\xi}]_\times)
\end{aligned} \qquad (12)$$

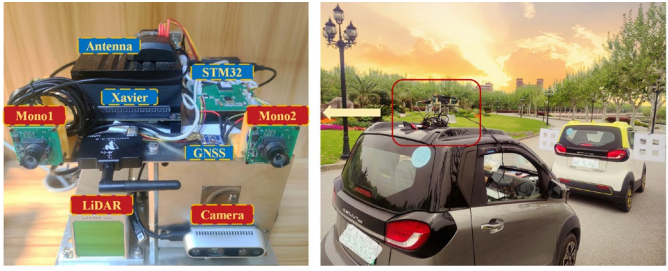

Figure 6. Multi-sensor suite and road test. Two calibration boards are placed in the front vehicle, and the suite is attached to the following car via suckers to collect data.

TABLE I. EXPERIMENTAL SETTINGS OF PARAMETERS

| Distance | $\delta_s$ | $\delta_{range}$ | $\delta_d^{Plane}$ | $\delta_d^{Line}$ |
|---|---|---|---|---|
| 3m | 0.20 | 0.2 | 0.05 | 0.01 |
| 5m | 0.15 | 0.4 | 0.05 | 0.015 |
| 8m | 0.10 | 0.5 | 0.06 | 0.02 |

thus the derivative of the residual $e$, defined as $\mathbf{P}_i^L - \pi(\mathbf{T}_L^C \mathbf{P}_i^L)$, to the perturbation is

$$\mathbf{J_T} = \frac{\partial e}{\partial \delta \boldsymbol{\xi}} = \lim_{\delta \boldsymbol{\xi} \to 0} \frac{e(\delta \boldsymbol{\xi} \oplus \boldsymbol{\xi}) - e(\boldsymbol{\xi})}{\delta \boldsymbol{\xi}} = \frac{\partial e}{\partial \mathbf{T}_L^C \mathbf{P}_i^L} \frac{\partial \mathbf{T}_L^C \mathbf{P}_i^L}{\partial \delta \boldsymbol{\xi}} \quad (13)$$

Here, the former term is

$$\frac{\partial e}{\partial \mathbf{T}_L^C \mathbf{P}_i^L} = \begin{bmatrix} \frac{\partial u_i}{\partial X_i} & \frac{\partial u_i}{\partial Y_i} & \frac{\partial u_i}{\partial Z_i} \\ \frac{\partial v_i}{\partial X_i} & \frac{\partial v_i}{\partial Y_i} & \frac{\partial v_i}{\partial Z_i} \end{bmatrix} = \begin{bmatrix} \frac{f_x}{Z_i} & 0 & -\frac{f_x X_i}{Z_i^2} \\ 0 & \frac{f_y}{Z_i} & -\frac{f_y Y_i}{Z_i^2} \end{bmatrix} \quad (14)$$

the latter term, by applying (12), is

$$\frac{\partial \mathbf{T}_L^C \mathbf{P}_i^L}{\partial \delta \boldsymbol{\xi}} \approx \lim_{\delta \boldsymbol{\xi} \to 0} \frac{(\mathbf{I} + [\delta \boldsymbol{\xi}]_\times) \exp([\boldsymbol{\xi}]_\times) \mathbf{P}_i^L - \exp([\boldsymbol{\xi}]_\times) \mathbf{P}_i^L}{\delta \boldsymbol{\xi}}$$
$$= \lim_{\delta \boldsymbol{\xi} \to 0} \frac{[\delta \boldsymbol{\xi}]_\times (\exp[\boldsymbol{\xi}]_\times \mathbf{P}_i^L)}{\delta \boldsymbol{\xi}} = \begin{bmatrix} \mathbf{I} & -[\mathbf{T}_L^C \mathbf{P}_i^L]_\times \\ \mathbf{0}^T & \mathbf{0}^T \end{bmatrix} \quad (15)$$

Now that the Jacobian $\mathbf{J_T}$ is obtained, we could feed it into an optimization backend such as Ceres, g2o, gtsam or SE Sync to solve an optimization problem as equation (10). Here we use Ceres.

IV. EXPERIMENTAL RESULTS

This section demonstrates the performance of the proposed approach in both qualitative and quantitative manners with comprehensive experiments. The robustness of our method is also verified by calibration scenes of different natures and various sensor configurations. By comparison with the state-of-the-art algorithm along this line, our method proves to produce a more accurate result.

A. Multi-Sensor Suite

The algorithm is deployed on an NVIDIA Xavier, which is also the ROS master responsible for all sensors' launches, and all the data will be recorded into its internal storage. We use the solid-state LiDAR Livox Mid-70 with both the horizontal and vertical FoV of 70.4° as the point cloud device, an Intel Realsense-D435i depth camera (resolution $1280 \times 720$), and multiple VEYE industrial monocular cameras (resolution $640 \times 540$), as shown in Figure 6.

The Pulse Per Second (PPS) signal is output by a GNSS receiver and used for LiDAR data synchronization, and an STM32 MCU is applied for the PPS frequency doubling to generate TTL signals that trigger multiple cameras. During data collection, the suite should be held static for several seconds to accumulate enough data.

B. Performance Comparison to Other Target-based Methods

To verify the accuracy and robustness of the proposed approach, both indoor and outdoor experiments are conducted. For the former, the calibration boards are placed on the ground, with a clearance of 10-50 cm to the background behind, or hung on the wall, as seen in Figure 4 (a) and Figure 5 (a). For the latter, our sensor suite is attached to a car using vacuum suction cups, and the calibration boards are suspended on a pole facing backwards at each side of the leading vehicle (as target), as depicted in Figure 6. Please note that a single board is enough in most cases if it could be placed stably and close enough to the sensor to fill the FoV.

The recommended experimental values of the thresholds are listed in Table I, according to the distance from the LiDAR to the calibration boards. The margin step $\delta_s$ determines the number of voxels, and is a user-determined parameter with

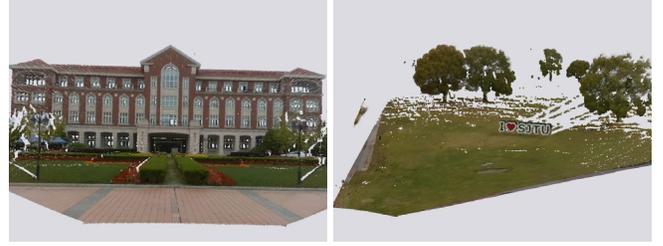

(a) Scene 1, distant building    (b) Scene 2, distant lawn

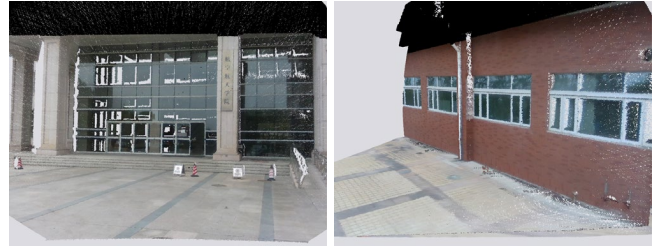

(c) Scene 3, close building    (d) Scene 4, close wall

Figure 7. Calibration scenes visualized by point cloud colorization.

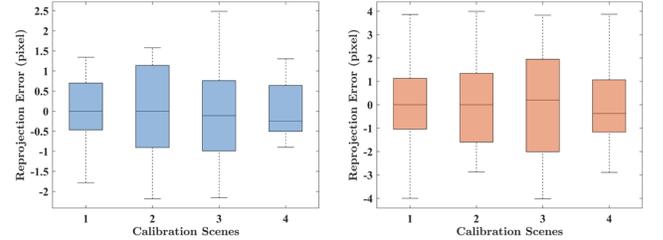

(a) Reprojection error of $U$ axis    (b) Reprojection error of $V$ axis

Figure 8. Reprojection errors of the image $u$ and $v$ axes by box plots, which summarize the statistical data, e.g., 25th percentile, median, 75th percentile.

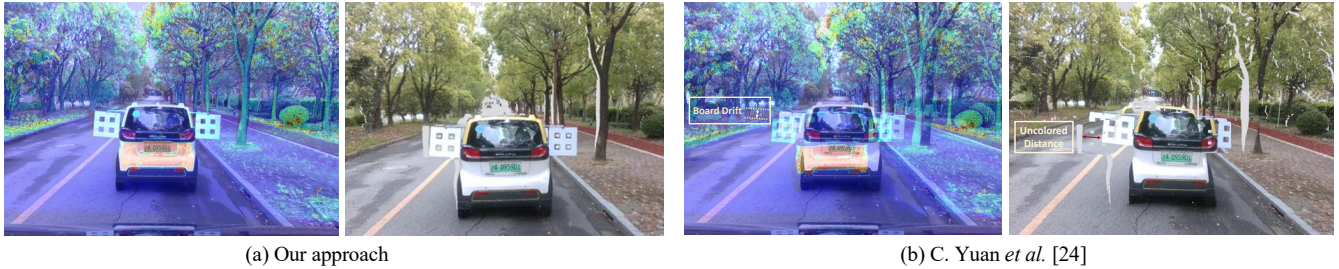

(a) Our approach            (b) C. Yuan *et al.* [24]

Figure 9. Comparison experiments with the state-of-the-art targetless method ([24]). For each method, the point cloud projection on image is shown on the left, while the colorized point cloud is visualized on the right. Both comparison shown that our method outperforms [24].

TABLE II. Performance Comparison with Other Target-Based Methods

| Approach | Mean | error < p(%) | | |
|---|---|---|---|---|
| | | $p=1$ | $p=5$ | $p=10$ |
| Cui [25] | 2.11 | **75.41** | 87.16 | 92.75 |
| Pandey [11] | 9.77 | 3.93 | 39.34 | 60.15 |
| **Ours** | **1.88** | 44.64 | **90.18** | **99.11** |

TABLE III. Performance Comparison with the Targetless Methods (Solid-State LiDAR)

| Criterion | Board Drift | | | Uncolored Distance | | |
|---|---|---|---|---|---|---|
| Scene | Distant | Close | Bad | Distant | Close | Bad |
| **Ours** | **1.41** | **5.10** | **3.61** | **0.02** | 0.09 | **0.06** |
| Yuan [24] | 7.07 | 6.50 | 44.65 | 0.05 | **0.06** | 0.34 |
| Manual | 12.72 | 8.94 | 8.13 | 0.21 | 0.16 | 0.09 |

Unit: Board drift measures pixel distance, uncolored distance is the 3D distance (m).

respect to the resolution, a decisive factor for the computational complexity; the margin point threshold $\delta_{range}$ is recommended to be large enough to decrease interaction of the laser beam with the background; the plane threshold $\delta_d^{Plane}$ and the line threshold $\delta_d^{Line}$ determine the thickness of the virtual plane and line; they should vary according to the reflectivity of the board and the size of the square hole.

The qualitative evaluation criterion evaluates the performance by exploiting point cloud colorization, which obtains the pixel RGB information of the corresponding 3D points with the estimated results, which was exemplified in Figure 1. We test our algorithm on multiple scenes with calibration boards attached to the leading car to calculate the extrinsic parameters, and then remove the leading vehicle and the calibration board to re-collect data in the same scenes. Figure 7 visualizes the calibration scenes via the above method, two of which with large depths and the other two with small depths. It can be seen that the pixel information converges to the real objects in each scene, which demonstrates both the viability and also, the robustness of the proposed approach.

The quantitative evaluation criterion focuses on the reprojection errors of the selected feature points, the result of which is presented as the boxplots shown in Figure 8. The majority of the 25th, 75th percentiles are within one pixel, validating the precision and accuracy of our method. For comparison, the normalized reprojection error in [25] is adopted to erase the scale effect, and the quantitative results of our method and previous target-based studies on the solid-state LiDAR are listed in Table II, where $error < p$ determines the percentage of points with the normalized reprojection error is less than $p$ pixels. It can be concluded from the table that, our method outperforms the others [11, 25] in terms of the mean value, and ours also boasts a small standard deviation, characterized by better performance for large $p$.

In brief, the quantitative criterion is the representation of the algorithm from the theoretical and data perspective, and the qualitative manner is a visual way complementary to the first, but is indispensable since the results that converge to a small reprojection error might plunge into a local minimum trap. Also note that, most implementations of previous target-based calibration approaches failed to detect corners, due to their heavy reliance on the sparse scanning pattern (compared to the relatively dense ROI of the solid-state LiDAR) of the mechanical LiDAR, as mentioned in [25]. Hence, in the following comparison, the target-based methods are not taken into account, and we compare our approach with the state-of-the-art targetless solid-state LiDAR calibration algorithm.

### C. Comparison with Targetless Method (Solid-State LiDAR)

To the authors' knowledge, among the plethora of targetless methods, only [24] are tailored to a solid-state LiDAR. Hence, the data collected in the four scenes in Figure 7 are evaluated in terms of the state-of-the-art method [24] and the proposed method. In addition, we add another road test that belongs to 'bad calibration scenes' mentioned in [24] (i.e., scenarios that lack sufficient edge features). The qualitative comparison of the additional scene (denoted as scene 5) is visualized in Figure 9, exploiting two common performance metrics, i.e. the point cloud projection and colorization. To enable a quantitative comparison, which is always neglected by the previous work, we propose two new metrics: 1) *board drift* $\delta_{bd}$, defined as the average translation of corners $\mathbf{p}_{original}^i$ in the original image and the overlaid point cloud $\mathbf{p}_{overlaid}^i$, derived from the projection method and 2) *uncolored distance* $\delta_{ud}$, defined as the maximum 3D distance of real position of the board edge $E_{board}$ and the displayed one $E_{display}$, derived from the colorization:

$$\delta_{bd} = \sum_{i=1}^{n} \frac{||\mathbf{p}_{original}^i - \mathbf{p}_{overlaid}^i||^2}{n} \quad (16)$$

$$\delta_{ud} = \max ||E_{board} - E_{display}||^2 \quad (17)$$

Both of the metrics are shown in Figure 9, and the results are summarized in Table III in terms of the three representative scenes, i.e. distant scene 1, close scene 3, and bad scene 5 of

Figure 7. We also offer a comparison to a mainstream manual calibration algorithm with that of the above two methods.

It is observed in Table III that our method and the manual one are robust in all three scenes, since they do not rely on the features extracted from the scenes, while [24] has poor performance in the lack-of-line-feature scene (like scene 5, shown in Figure 9), especially in the 'road and tree' environment lacking stable structural features and easily interfered by dynamic disturbance like wind. From the perspective of accuracy, the targetless method is slightly inferior or even on par with the target-based approach in the close scene where plentiful observations of the building provide favorable features. However, in the distant scene, due to the unevenly-distributed characteristics of the point cloud, some far objects are not observed adequately (this can be seen in Figure 7 (a), take the edges and contours of the building for examples), thus increasing difficulties in feature detection of their method and exerting inferior results. The manual method has mediocre accuracy and generalization capacity compared with the others.

In brief, though the targetless method [24] brings convenience in calibration, our method has higher accuracy and is more robust, which is suitable for applications regarding strict demands. The targetless method remains a feasible and handy way in feature-rich scenes while the proposed method performs better in feature-barren scenes.

## V. CONCLUSION

This letter proposed a robust self-calibration approach for camera and solid-state LiDAR. We firstly re-examined the characteristic of the point cloud from the unique prism-spinning solid-state LiDAR, and found out that the vacant points, defined as depth discontinuous measurements, are the main barriers of the conventional depth discrimination methods. To address this problem, we designed a removal procedure using voxelization and depth-discontinuous points upon the tailored perforated calibration boards which in the feature-barren scenes, retains more valid scans than the state-of-the-art method that used only depth-continuous points. The custom-designed calibration target is also recognizable by the overlapping camera and we propose two pixel constraints, called rectangularity and circularity. Comparison experiments with the target-based counterparts demonstrate the favorable precision of our approach from the qualitative and the quantitative perspectives, and comparison with the state-of-the-art targetless method for the solid-state LiDAR concerning multiple scenes depict the robustness of our method, especially in open field scenes.